\documentclass[smallextended]{svjour3}  
\smartqed  

\usepackage{xcolor,framed} 

\colorlet{shadecolor}{yellow}
\usepackage[pdftex]{graphicx}
\graphicspath{{../pdf/}{../jpeg/}}
\DeclareGraphicsExtensions{.pdf,.jpeg,.png}

\usepackage[]{algorithm2e}
\SetArgSty{textup}

\usepackage[normalem]{ulem}
\usepackage[american]{babel}
\usepackage[numbers]{natbib}
\usepackage[cmex10]{amsmath}
\usepackage{array}
\usepackage{mdwmath}
\usepackage{mdwtab}
\usepackage{eqparbox}
\usepackage{url}
\usepackage{smartdiagram}
\usepackage{verbatim}
\usepackage{subcaption} 

\usepackage{graphicx} 
\journalname{}
\begin{document}

\title{Playing to distraction: towards a robust training of CNN classifiers through visual explanation techniques \thanks{Partial financial support was received from HAT.tec GmbH. This work has been financially supported in part by European Union ERDF funds, by the Spanish Ministry of Science and Innovation (research project PID2019-109238GB-C21), and by the Principado de Asturias Regional Government (research project IDI-2018-000176). The funders had no role in the study design, data collection, analysis, and preparation of the manuscript.}}


\titlerunning{Playing to distraction: towards a robust training of CNN classifiers}   

\author{David Morales \and Estefania Talavera \and Beatriz Remeseiro 
}


\institute{David Morales \at Andalusian Research Institute in Data Science and Computational Intelligence, University of Granada, 18071 Granada, Spain. 
\at{HAT.tec GmbH. c/o Universität der Bundeswehr,  Werner-Heisenberg-Weg 39, 85579 Neubiberg, Germany.}\\
\email{david.morales@hattec.de}
\and
Estefania Talavera \at Department of Computer Science, University of Groningen. Nijenborgh 9, 9747 AG Groningen, Netherlands. \\
\email{e.talavera.martinez@rug.nl}
\and Beatriz Remeseiro \at Department of Computer Science, Universidad de Oviedo. Campus de Gij\'on s/n, 33203 Gij\'on, Spain. \\
\email{bremeseiro@uniovi.es}
}

\date{Received: date / Accepted: date}

\maketitle
\begin{abstract}
The field of deep learning is evolving in different directions, with still the need for more efficient training strategies. In this work, we present a novel and robust training scheme that integrates visual explanation techniques in the learning process. Unlike the attention mechanisms that focus on the relevant parts of images, we aim to improve the robustness of the model by making it pay attention to other regions as well. Broadly speaking, the idea is to \textit{distract} the classifier in the learning process by forcing it to focus not only on relevant regions but also on those that, \textit{a priori}, are not so informative for the discrimination of the class. We tested the proposed approach by embedding it into the learning process of a convolutional neural network for the analysis and classification of two well-known datasets, namely Stanford cars and FGVC-Aircraft. Furthermore, we evaluated our model on a real-case scenario for the classification of egocentric images, allowing us to obtain relevant information about peoples' lifestyles. In particular, we work on the challenging EgoFoodPlaces dataset, achieving state-of-the-art results with a lower level of complexity. The results obtained indicate the suitability of our proposed training scheme for image classification, improving the robustness of the final model.

\end{abstract}


\section{Introduction}

Nowadays, the potential of convolutional deep learning models for the task of image classification has been proven. Research in this field has followed different directions namely, new architecture and framework proposals \cite{richards2019deep, khan2019novel}, training methods \cite{wu2017effective, xu2018semantic}, multi-tasking \cite{zhang2020deep, luvizon20182d}, 
attention mechanisms \cite{li2019understanding, jain2020deep}, explainability and interpretability \cite{samek2017explainable, vellido2019importance}, among others.


New techniques such as attention mechanisms allow to force the model to pay attention to certain features, whilst explainable artificial intelligence techniques allow to interpret the model and know what is happening during the learning process. However, to the best of our knowledge, the combination of both approaches has not been explored. Inspired by this lack of combination, we aim to improve the training procedure by interpreting the model and focusing it on certain regions of interest. 
To this end, our proposed approach is based on modifying the classical training procedure to include online information and thus adapt the learning process based on the features on which the network is focused.

More specifically, we propose a new training scheme that benefits from the saliency maps provided by visual explanation techniques. Our hypothesis is that, by the end of the training phase, the model should use as many features as possible to make a robust prediction. In this sense, we apply a visual explanation algorithm to identify the regions on which the model bases its decisions. After identifying those relevant areas, we partially occlude them trying to \textit{distract} the model in some way and forcing the detection of other regions that, a priori, are weak (i.e., not so informative for the discrimination of the class). Our intention is to highlight that the model should not forget what the occluded regions mean, but it should learn to recognize other features to make a decision. This is ensured as the occluded images are combined with the original ones during the learning process.

We think fine-grained image classification problems could benefit the most from this approach, as they have many classes that differ from each other in small details, and our training approach forces the network to find them. For this reason, we evaluated the proposed training scheme on two well-know datasets namely Stanford cars \cite{stanfordCars} and FGVC-Aircraft \cite{aircraftDataset}, composed of 16,185 and 10,000 images respectively, and used in fine-grained recognition. In addition, we carried out some experiments on top of different backbone architectures to demonstrate that our proposal improves the performance regardless of the respective network.

Furthermore, we evaluate the robustness of our model in a real-scenario case study: recognizing the food-related scene that an egocentric image depicts. The analysis of egocentric images is an emerging field within computer vision that has gained attention in recent years \cite{damen2018scaling}. Images captured by wearable cameras during daily life allow recording information about the lifestyle of the users from a first-person  perspective \cite{bolanos2016toward, talavera_dataset}. The analysis of this information can be used to improve peoples' health-related habits \cite{gelonch2020effects}. In particular, the analysis of food-related egocentric images can be a powerful tool to analyze peoples' nutritional habits, being the focus of previous research \cite{macnet,talavera_dataset}. In this context, we carried out some experiments on the EgoFoodPlaces dataset \cite{talavera_dataset}, which is composed of 33,801 images and describes food-related locations gathered by 11 camera wearers throughout their daily life activities.

The contributions of this research work are three-fold:
\begin{enumerate}
    \item A novel training scheme for CNN image classification that makes use of visual explanation techniques, with the main aim of improving the robustness and the generalization ability of the trained models.
    \item The experiments carried out demonstrate the competitiveness of our training scheme, which outperforms the classical approach on two public datasets commonly used in fine-grained recognition tasks, regardless of the backbone architecture.
    \item Our proposed method achieves competitive results in a real-case scenario that addresses the classification of egocentric photo-streams depicting food-related scenes.
\end{enumerate}

The rest of the paper is organized as follows. Section \ref{section:related_work} includes an overview of related works. Section \ref{section:proposed_approach} presents the proposed training approach. Section \ref{section:experimental_results} introduces the two datasets for fine-grained recognition, describes the experiments carried out and analyzes the obtained results. Section \ref{section:case_study} describes and evaluates the case study focused on egocentric vision. Finally, Section \ref{section:conclusion} closes with our conclusions and future lines of research.

\section{Related Work}\label{section:related_work}

While the very first machine learning systems were easily interpretable, the last years have been characterized by an upsurge of opaque decision systems, such as deep neural networks (DNNs) \cite{xai_herrera,xai_natalia}. DNNs are the state-of-the-art on many machine learning tasks due to their great generalization and prediction skills. However, they are considered \textit{black-box} machine learning models. In this context, there has been a growing influx of work on explainable artificial intelligence. Post-hoc local explanations, which refer to the use of interpretation methods after training a model, and feature relevance methods are increasingly the most adopted approaches to explain DNNs \cite{xai_herrera}. In this section, we review some methods that produce \textit{visual explanations} for decisions of a large class of DNN-based models, making them more transparent and reliable. 

Most of these visual explanation techniques provide heat maps to identify the regions of the input images that networks look at when making predictions, allowing the data to be interpreted at a glance. Note that these heat maps are also referred to in the literature as sensitivity maps, saliency maps, or class activation maps. Class activation mapping (CAM) \cite{cam} is a well-known procedure for generating class activation maps using global average pooling in CNNs. Their authors expect each unit to be activated by some visual pattern within its receptive field. The class activation map is nothing more than a weighted linear sum of the presence of these visual patterns at different spatial locations. By simply upsampling the class activation map to the size of the input image, 
they can analyze the most relevant image regions to identify the particular category.
However, CAM can only be used with a restricted set of layers and architectures.

A class-discriminative localization technique called gradient-weighted class activation mapping (Grad-CAM) was proposed in \cite{gradcam}. In fact, it is a generalization of CAM that can be applied to a significantly broader range of CNN families. Grad-CAM uses the gradients of any target concept flowing into the final convolutional layer to produce a coarse localization map, highlighting the regions of the image that are relevant for the prediction. Given an image and a class of interest (e.g., \textit{tiger cat}) as inputs, Grad-CAM forward propagates the image through the convolutional part of the model and then through task-specific computations to obtain a raw score for the category. The gradients are set to 0 for all classes except for the desired class (\textit{tiger cat}), which is set to 1. This signal is then backpropagated to the rectified convolutional feature maps of interest, which are combined to compute the coarse Grad-CAM localization that represents where the model looks at to make the corresponding decision. Finally, they point-wise multiply the heat map with guided backpropagation, thus obtaining also guided Grad-CAM visualizations, which are both high-resolution and concept-specific.

Another visual explanation method was presented in \cite{occlusion}, in which input images are perturbed by occluding all their patches, in an iterative process, and classifying the occluded images. This idea allows the authors to analyze how the top feature maps and the classifier output change, revealing structures within each patch that stimulate a particular feature map. However, the use of this method requires generating multiple occluded samples and their classification, making it computationally expensive.

Ribeiro et al.~\cite{lime} proposed the local interpretable model-agnostic explanations (LIME) technique, which allows to explain the predictions of any classifier in an interpretable and faithful manner. Given the original representation of the instance being explained, they get new samples by perturbing the original representation. They use those samples to approximate the classifier with an interpretable model. Just as the method above, the use of multiple samples implies to apply the classifier several times given one instance.

Some of these visual explanation techniques generate noisy sensitivity maps. In this context, Smilkov et al.~\cite{smoothgrad} proposed SmoothGrad, a technique to reduce the noise in the sensitivity maps produced by visual explanation techniques based on gradients. Their idea was to sample images similar to the original ones by adding some noise. Then, they produced intermediate sensitivity maps for each image and took the average of them as the final sensitivity map.

Finally, it is worth highlighting some applications of the saliency maps generated by visual explanation techniques. Sch\"ottl~\cite{schottl2020light} used Grad-CAM maps to improve the explainability of classification networks. More specifically, the idea was to introduce some measures obtained from the Grad-CAM maps in the loss function. Cancela et al.~\cite{cancela2020scalable} proposed a saliency-based feature selection method that selects the features that contain a higher discrimination result, allowing to provide robust and explainable predictions in both classification and regression problems.

\subsection{Egocentric photo-streams}

Following, we review some recent works on 
egocentric photo-streams, mainly focused on the classification of food-related scenes, such as our case study.

Egocentric image analysis is a field within computer vision related to the design and development of algorithms to analyze and understand photo-streams captured by wearable cameras \cite{talavera_dataset}. These cameras are capable of capturing images that record visual information of our daily life, known as \textit{visual lifelogging}, to create a visual diary with activities of first-person life. The analysis of these egocentric photo-streams can improve peoples' lifestyle by analyzing social patterns \cite{herruzo2017analyzing}, social interactions \cite{aghaei2016whom}, or food behavior \cite{talavera2020eating}. 

In recent years, there is a growing interest in egocentric photo-streams giving their potential for assisted living. For instance, Furnari et al. \cite{furnari1} presented a benchmark dataset containing egocentric videos of eight personal locations and proposed a multi-class classifier to reject locations not belonging to any of the categories of interest for the end-user.

As for food-related scene recognition, Sarker et al. \cite{macnet} addressed this task by proposing a multi-scale atrous CNN \cite{atrous} to analyze lifelogging images and determine people's recurrences in food places throughout their day. Later, Talavera et al. \cite{talavera_dataset} presented the EgoFoodPlaces dataset, composed of more than 33,000 images organized in 15 food-related scene classes. This dataset was recorded by 11 users while spending time on the acquisition, preparation, or consumption of food. The dataset was manually labeled into a total of 15 different food-related scene classes like \textit{bakery shop}, \textit{bar}, or \textit{kitchen}. Taking into account the relation of the studied classes, a taxonomy for food-related scene recognition was introduced. Furthermore, the authors proposed a hierarchical classification model based on the aggregation of six VGG16 networks \cite{vgg} over different subgroups of classes, emulating the proposed taxonomy. This is, to the best of our knowledge, the state-of-the-art in the recognition of food-related scenes in egocentric images.


\section{Methodology} \label{section:proposed_approach}

\begin{figure}[htb]
    \includegraphics[width=\textwidth]{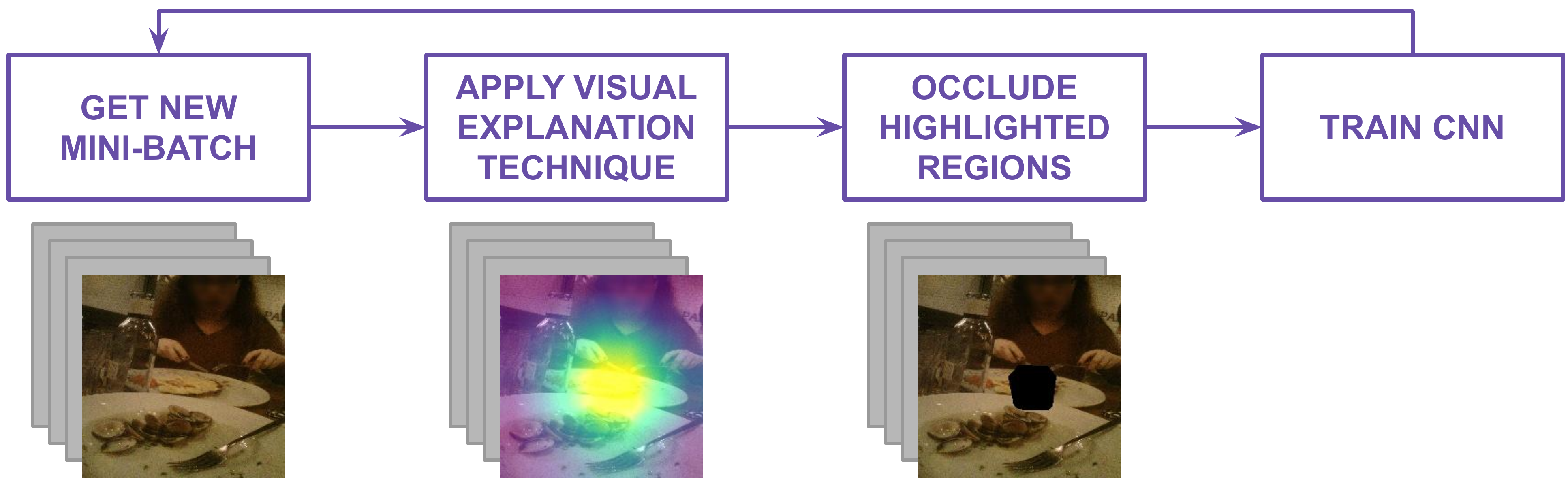}
    \caption{Workflow of our alternative training scheme, which (1) gets a new mini-batch of input images, (2) applies a visual explanation technique to generate the heat maps, (3) occludes the regions highlighted in the previous step, and (4) trains the CNN classifier.} \label{fig:workflow}
\end{figure}

We propose a novel training approach to improve the robustness of CNNs in image classification. Figure \ref{fig:workflow} illustrates the different steps of the proposed scheme, which are subsequently explained in depth.

Let consider the classical mini-batch gradient descent \cite{dekel2012optimal} training algorithm where, on each training step, the mini-batch is first fed into the neural network, then the gradient is computed, and finally, the calculated gradient is used to update the weights of the network. We propose to modify the training step to apply the new scheme over each mini-batch with a probability $p \in (0,1)$; i.e., with a probability $1-p$, the images in the mini-batch kept unchanged and the classical training step is performed as usual. Note that the probability $p$ belongs to the open interval $(0,1)$. 
$p = 0$ would mean that our training scheme is not applied (i.e., the classical training procedure is used instead). $p = 1$ would mean that only the modified images are used, 
making model convergence difficult. Preliminary experimentation suggests applying the method with values of $p \leq 0.5$ to guarantee that both occluded and original images are used in the learning process. Therefore, with a probability $p \in (0,1)$, our training scheme is applied as follows: 

\begin{enumerate}
    \item Using the current weights of the network, we do inference over the current mini-batch and apply a visual explanation method to get a heat map for each image in the mini-batch. These heat maps highlight the regions where the current model focuses its attention to classify the corresponding image.
    \item After that, we occlude the areas corresponding to those highlighted regions, forcing the model to look at other regions in the image. For each image in the mini-batch, we normalize its heat map and get a weight $w \in [0,1]$ for each pixel. Next, we select all the pixels whose weight $w$ is over a threshold $th$. The selected pixels are erased by setting them to 0, calling this approach 0-occlusion. As a result, we obtain the occluded images of the mini-batch.
    \item Finally, we train our model making use of the occluded mini-batch.
\end{enumerate}

Algorithm \ref{alg:pseudoCodigo} shows the pseudo-code of our proposed training method according to the 0-occlusion approach. Note that once the mini-batch is modified, the training step continues as usual (i.e., the gradient is calculated and the weights are updated). We think it is important to highlight that the model should not forget what the occluded regions mean, but learn to recognize other parts of the image to make a decision. This is guaranteed as the occluded images are used 
only for some mini-batches, according to the $p$ hyper-parameter, while the original ones are used for the rest of them. 

\begin{algorithm}[htb]
 \KwData{trainingSet, model, $p$, $th$}
 \KwResult{the model trained using the proposed approach}
 \For{miniBatch $in$ trainingSet}{
    $r$ = random(0,1)\;
    \If{$r \leq p$}{
    \For{(image, label) $in$ miniBatch}{
        heatMap = visualExplanation(image, label, model, lastConvLayer)\;
        heatMap = minMaxNorm(heatMap)\;
        selectedPixels = $[$heatMap $> th]$\; 
        image$[$selectedPixels$]$ = 0\;
        }
    }
    train(model, miniBatch)\;
 }
 \caption{Pseudo-code of the proposed training scheme using 0-occlusion.} \label{alg:pseudoCodigo}
\end{algorithm}

The proposed approach is compatible with any of the visual explanation methods presented in Section \ref{section:related_work} and, in general, with any method that generates a heat map to explain the decision of a CNN for a given target image. Among all these techniques, we choose Grad-CAM \cite{gradcam} because it uses the flow of the gradients from the last convolutional layer to compute the heat maps, making it computationally less expensive than other methods like LIME \cite{lime} or SmoothGrad \cite{smoothgrad}. These other techniques apply inference several times on images generated by perturbing the target image to compute the heat maps. In other words, Grad-CAM does inference once per image while other techniques do inference several times per image, which makes the former more appropriate for the problem at hand.

Summarizing, the heat maps provided by Grad-CAM highlight the relevant regions of the image for predicting the ground truth class. By occluding them, the model is forced to look at other regions to make the decision. The initial regions should not be forgotten by the model, but other parts of the images should also be taken into account in the learning process. In this manner, the model improves its robustness and generalization capabilities.

\section{Experimental framework and results}\label{section:experimental_results}

In this section, we present two datasets used to evaluate the proposed method. Next, we describe the implementation details as well as the two experiments carried out, including the evaluation metrics considered. Finally, we report and analyze the results obtained in both experiments: (1) a comparison between the proposed method and some  variants of it, and (2) a comparison with standardized baselines.

\subsection{Datasets} \label{sec:datasets}

We evaluated our proposed method on two well-known datasets: the Stanford cars dataset \cite{stanfordCars}, and the fine-grained visual classification of aircraft (FGVC-Aircraft) benchmark dataset \cite{aircraftDataset}. Both datasets were used as part of the fine-grained recognition challenge FGComp 2013, which ran jointly with the ImageNet Challenge 2013\footnote{http://image-net.org/challenges/LSVRC/2013/}.

The Stanford cars dataset contains 16,185 images of 196 car models covering sedans, SUVs, coupes, convertibles, pickups, hatchbacks, and station wagons; and it is officially split into 8,144 training and 8,041 test images. The FGVC-Aircraft dataset contains 10,000 images of aircraft, with 100 images for each of 100 different aircraft model variants; and it is officially split into 6,667 training and 3,333 test images. 

\subsection{Implementation details} \label{sec:exp_implementation}

The techniques and parameters used for experimentation are explained in the following. 
We used the Adam optimization algorithm \cite{adam} with the following parameters: learning rate $\alpha=0.00005$, $\beta_1=0.9$, $\beta_2=0.999$, and $\epsilon=0.0000007$. Regarding the training step, we used a batch size of 16 and the images were resized to $224 \times 224$. The outputs were monitored using the validation accuracy to apply an early stopping strategy, based on which the training process finished after 30 epochs with no improvement. Additionally, we applied the following classical data augmentation techniques: horizontal flip, rotation $[-40º, 40º]$, random channel shift $[-30, 30]$, and image brightness change $[0.5, 1.5]$. 

The proposed method was implemented on TensorFlow \cite{tensorflow} and Keras \cite{keras}, and the code is available for download\footnote{https://github.com/DavidMrd/Playing-to-distraction}. Starting from the training algorithm provided in Keras, we modified the training step to apply our method over each mini-batch with a probability $p$, as described in Section \ref{section:proposed_approach}. According to some preliminary experiments, we applied the proposed method with a probability $p=0.25$, and the threshold for the occlusion step was set to $th=0.85$. 

\subsection{Experimental setup} \label{sec:exp_setup}

This section describes the two experiments designed to evaluate our training scheme. Both experiments were applied to each dataset individually and compared with other approaches.
As for the experimentation itself, we kept the original split in training and test sets for the two considered datasets (see Section \ref{sec:datasets}). For validation purposes, we randomly divided the original training dataset into two parts: 75\% training and 25\% validation. Then, we trained the model and evaluated it on the isolated test set, using the performance metrics 
described in Section \ref{sec:exp_evaluation}. This validation procedure was repeated five 
times. We report the average performance and the standard deviation calculated across the five runs.

\subsubsection{Experiment 1}

The objective of this experiment is to test several setups of our training scheme and compare them with a baseline. In particular, we used a ResNet50 \cite{resnet}, a very popular network successfully applied to different image classification tasks. The different configurations are detailed as follows:

\begin{enumerate}

\item  \textbf{Baseline.} In order to compare our method with a baseline, we trained a ResNet50 using the classical training approach (i.e., without applying the proposed method). We called this model fine-tuned ResNet50 (FT-ResNet50) because it is a model pre-trained on the ImageNet dataset \cite{imagenet}, whose parameters were fine-tuned with the corresponding dataset.

\item \textbf{Our approach.} We trained a ResNet50 using the proposed training method, which is based on Grad-CAM visualizations and illustrated in Figure \ref{fig:workflow}. More specifically, we used the weights from the ResNet50 model pre-trained on ImageNet \cite{imagenet}, and then we fine-tuned them using the corresponding dataset and our training scheme. Note that, during the learning process, the Grad-CAM algorithm was applied to the last convolutional layer of the ResNet50, as indicated in \cite{gradcam}.

\item \textbf{Other setups.} Aiming at demonstrating the adequacy of the 0-occlusion approach, we also conducted some experiments in which the pixels were set to a random value (R-occlusion) and 1 (1-occlusion). 
\end{enumerate}

\subsubsection{Experiment 2}

This experiment aims to demonstrate the adequacy of our training scheme regardless of the backbone architecture considered. In this sense, we applied it to two well-known backbone architectures, different from ResNet50, using the following configurations:

\begin{enumerate}
    \item \textbf{Baselines.} We trained two architectures commonly used in the literature, InceptionV3 \cite{inceptionV3} and DenseNet \cite{huang2017densely}, using the classical approach. We called them FT-InceptionV3 and FT-DenseNet, respectively, because they were pre-trained on ImageNet and fine-tuned with the corresponding dataset.
    \item \textbf{Our approach.} We trained the two backbone architectures considered, InceptionV3 and DenseNet, using the proposed training scheme (see Figure \ref{fig:workflow}). As in the previous experiment, we used the weights from these two architectures pre-trained on ImageNet, and then we fine-tuned them with the corresponding dataset and our training scheme. Regarding the Grad-CAM algorithm, it was applied to the last convolutional layer of the networks as described in \cite{gradcam}. 
    
\end{enumerate}

\subsection{Evaluation} \label{sec:exp_evaluation}

In order to evaluate the performance of the proposed models and make a fair comparison with other approaches, we computed some popular metrics in image classification tasks: accuracy, precision, recall, and F-score (F1). These metrics are defined as follows:

\begin{equation}
    Accuracy = \frac{TP+TN}{TP+TN+FP+FN}
\end{equation}

\begin{equation}
    Precision = \frac{TP}{TP+FP}
\end{equation}

\begin{equation}
    Recall = \frac{TP}{TP+FN}
\end{equation}

\begin{equation}
    F1 = 2*\frac{Precision*Recall}{Precision+Recall}
\end{equation}
where $TP$, $FP$, $TN$, and $FN$ stand for true positives, false positives, true negatives, and false negatives, respectively.

\subsection{Results}

In this section, we report and analyze the results obtained in the two experiments described above.

\subsubsection{Experiment 1}

Table \ref{tab:resultados_resnet} shows the results obtained for the different configurations. As can be observed, our training scheme provides very competitive results regardless of the setup used for the occlusion. Analyzing the four metrics considered, the three setups outperform the baseline method (FT-ResNet50), which was trained with the classical learning procedure, in both datasets. 
Focusing on our proposal (0-occlusion), it achieves a gain of more than 2 percent in the Standford cars dataset and about 2 percent in the FGVC-Aircraft dataset. In order to demonstrate the relevance of this improvement, we applied a statistical t-test that allows us to determine if there is a significant difference between the baseline (FT-ResNet50) and our proposal (0-occlusion). Notice that we used a paired sample, two-tailed t-test. As a result, we can confirm that our proposal significantly outperforms the baseline in terms of accuracy, with a significance level of 0.05.

If we analyze the behavior of the three different setups considered for the proposed training scheme, we can see that both 0-occlusion and 1-occlusion provide better results than R-occlusion, with a very slight difference in favor of the former (0-occlusion). The experiments show that, when using random values for the occlusion procedure, the model does not benefit so much from the \textit{distraction} applied to the model, by forcing it to look at new regions in the input images. This behavior is discussed in detail below, with some qualitative results that aim at illustrating the impact of the proposed method.
 
\begin{table*}[htb]
    \centering
    \begin{tabular}{l|c|c|c|c}
    \hline  
    \multicolumn{5}{c}{Stanford cars} \\ \hline 
    & FT-ResNet50 & 0-occlusion & R-occlusion & 1-occlusion \\ \hline 
     
    Accuracy & $0.849\pm 0.009$ & \boldmath$0.871\pm0.007$ & $0.860\pm0.009$  & $0.869 \pm 0.008$ \\	 		
    Precision & $0.855 \pm0.007 $& \boldmath$0.876\pm0.007$ & $0.866\pm0.008$ & $0.873 \pm 0.008$ \\ 
    Recall & $0.849\pm0.009$ & \boldmath$0.870\pm0.008$ & $0.860\pm0.009$ & $0.868 \pm 0.009 $ \\
    F1 & $0.848\pm0.009$ & \boldmath$0.870\pm0.008$ & $0.859\pm0.009$ & $0.867\pm0.009$ \\ \hline \hline
    
    \multicolumn{5}{c}{FGVC-Aircraft} \\ \hline   
    & FT-ResNet50 & 0-occlusion & R-occlusion & 1-occlusion \\ \hline 
     
    Accuracy & $0.731 \pm 0.013$& \boldmath$0.749 \pm 0.005$ & $0.739 \pm 0.012$ & $0.743 \pm 0.005$ \\
    Precision & $0.746 \pm 0.011$ & \boldmath$0.762 \pm 0.005 $ & $0.755 \pm 0.010$ & $0.759 \pm 0.004$ \\ 
    Recall & $0.731 \pm 0.013 $& \boldmath$0.749 \pm 0.005$ & $0.739 \pm 0.012$& $0.743 \pm 0.005 $ \\ 
    F1 & $0.731 \pm 0.014$& \boldmath$0.748 \pm 0.005$ & $0.739 \pm 0.012$& $0.743 \pm 0.005 $\\ \hline  
    \end{tabular}
        
    \caption{Classification performance, averaged across five runs, of the different approaches on the Stanford cars \cite{stanfordCars} and FGVC-Aircraft \cite{aircraftDataset} datasets. Best results are in bold.}
    \label{tab:resultados_resnet}
\end{table*}

\begin{figure} 
    \centering
    \begin{tabular}{c@{\hskip 0.01in}c@{\hskip 0.01in}c@{\hskip 0.01in}c@{\hskip 0.01in}c@{\hskip 0.01in}}
   
     \begin{subfigure}[b]{0.20\textwidth}
  \includegraphics[width=2.0cm, height=2.0cm]{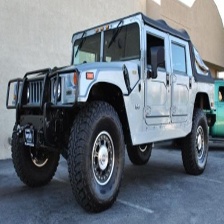}
     \end{subfigure}
     &
     \begin{subfigure}[b]{0.20\textwidth}
  \includegraphics[width=2.0cm, height=2.0cm]{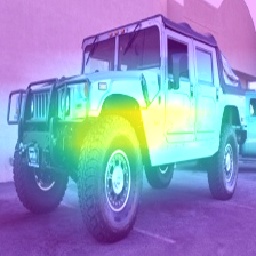}
     \end{subfigure}
     &
     \begin{subfigure}[b]{0.20\textwidth}
  \includegraphics[width=2.0cm, height=2.0cm]{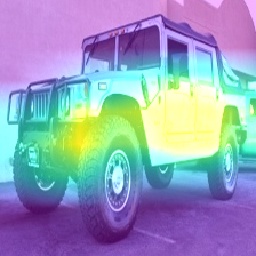}
     \end{subfigure}
    &
    \begin{subfigure}[b]{0.20\textwidth}
  \includegraphics[width=2.0cm, height=2.0cm]{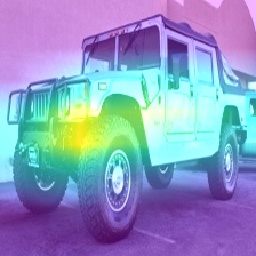}
     \end{subfigure}
    &
   \begin{subfigure}[b]{0.20\textwidth}
  \includegraphics[width=2.0cm, height=2.0cm]{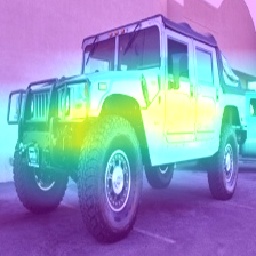}
   \end{subfigure}
  \\
  
       \begin{subfigure}[b]{0.20\textwidth}
  \includegraphics[width=2.0cm, height=2.0cm]{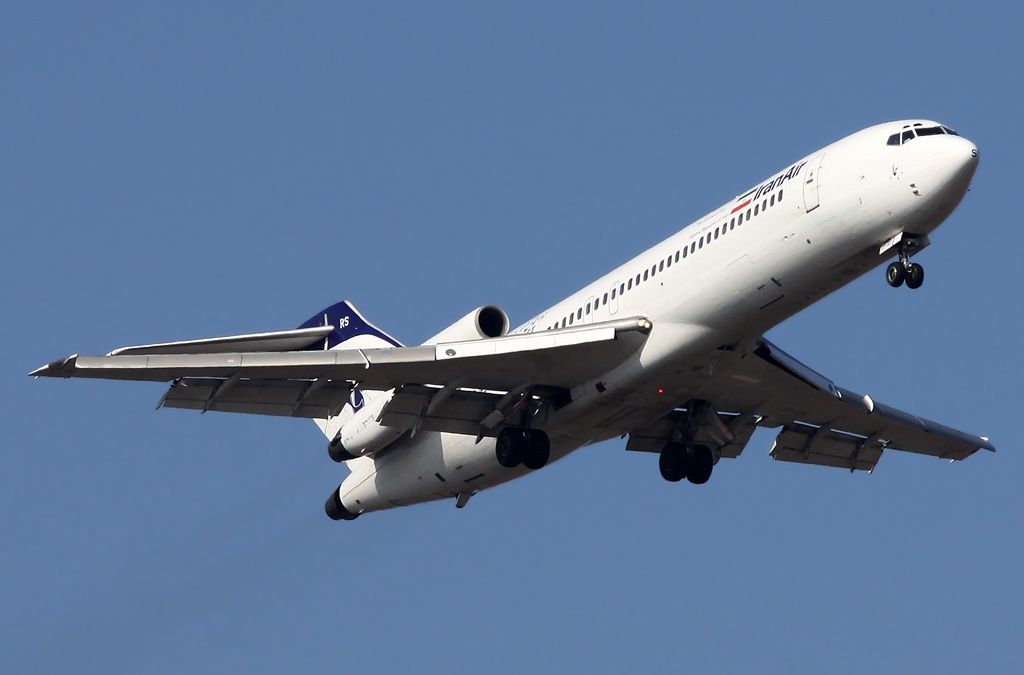} 
  \caption{}
     \end{subfigure} 
     &
     \begin{subfigure}[b]{0.20\textwidth}
  \includegraphics[width=2.0cm, height=2.0cm]{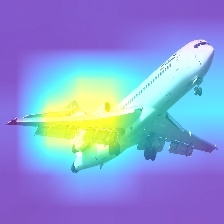}
    \caption{}
     \end{subfigure}
     &
     \begin{subfigure}[b]{0.20\textwidth}
  \includegraphics[width=2.0cm, height=2.0cm]{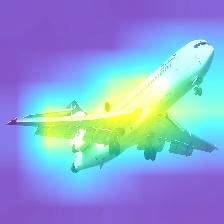}
   \caption{}
     \end{subfigure}
    &
    \begin{subfigure}[b]{0.20\textwidth}
  \includegraphics[width=2.0cm, height=2.0cm]{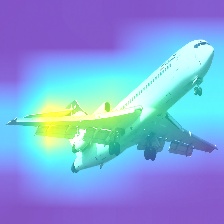}
   \caption{}
     \end{subfigure}
    &
   \begin{subfigure}[b]{0.20\textwidth}
  \includegraphics[width=2.0cm, height=2.0cm]{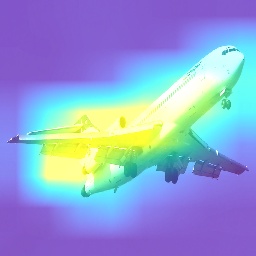}
      \caption{}
     \end{subfigure}
    \end{tabular}
    
    \caption{(a) Input images from the Stanford cars (top) and FGVC-Aircraft (bottom) datasets, (b) heat maps generated by Grad-CAM for the baseline FT-ResNet50, and heat maps generated by Grad-CAM for the model trained with the proposed training scheme using (c) 0-occlusion, (d) R-occlusion, and (e) 1-occlusion.}
    \label{fig:analysis}
\end{figure}

Figure \ref{fig:analysis} depicts two representative images of the two datasets used for experimentation, Stanford cars and FGVC-Aircraft, along with the heat maps generated by Grad-CAM for the different methods analyzed: the baseline FT-ResNet50 and the three setups for the proposed training approach. As can be observed, the models trained with the proposed approach, regardless of the setup, base their decisions on more features than the one trained using a classical approach (FT-ResNet50). While the baseline method seems to base its decisions just on a local area of the image, the models trained with the proposed approach seem to react to almost the whole object. Analyzing the different configurations, we can see that both 0-occlusion and 1-occlusion show a similar behavior, reacting to the whole object, which explains the achieved results in both cases. However, the R-occlusion version behaves differently since it reacts to many features but with a low level of confidence. That is, occluding the selected pixels with a fixed value (0 or 1) allows us to achieve better results than occluding the relevant regions with a random value. The reason for this behavior could be that, when using a fixed value, the model learns to ignore these areas and looks at other regions, whereas the model does not benefit as much from this idea when using a different value each time. It is worth noting that using 0-occlusion is somewhat similar to the well-known dropout \cite{dropout}, a regularization technique in which some connections are disabled during the learning phase. This would explain why this approach gets slightly better results than the 1-occlusion version.

%

Finally, Table \ref{tab:epochs} shows the number of epochs and the seconds per epoch needed to train the baseline (FT-ResNet50) and our proposal (0-occlusion). As can be observed, our training scheme requires more computational time per epoch and more epochs to converge than the classical procedure. Regarding the increment in terms of seconds per epoch, it is lower than $19\%$. Note that this time only depends on the image resolution and the hardware, so it is the same for both datasets. With respect to the increment in the number of epochs, it is $\approx32\%$ for the Stanford cars dataset and $\approx53\%$ for the FGVC-Aircraft dataset. Nevertheless, it is worth noting that, for application purposes, this computational time is not decisive since the training procedure is carried out only once before the model is put into production, after defining its architecture and setting its hyper-parameters. As our method is applied during the learning process, the computation time in the test phase is not affected.

\begin{table*}[htb]
    \centering
    \begin{tabular}{l|cc|cc}
    \cline{2-5}  
    & \multicolumn{2}{c|}{FT-ResNet50} & \multicolumn{2}{c}{0-occlusion} \\ 
    & Standford cars & FGVC-Aircraft  & Standford cars & FGVC-Aircraft \\ \hline 
    Number of epochs & $98.8 \pm 6.78$ & $113.4 \pm 10.97$ & $130 \pm 10.38$ & $174 \pm 13.87$\\ 
    Seconds per epoch* & \multicolumn{2}{c|}{$153 \pm 0.00$} & \multicolumn{2}{c}{$175 \pm 0.00$} \\ \hline
    \multicolumn{5}{l}{* Network input size: $224 \times 224 \times 3$. Hardware: NVIDIA T4 Tensor Core GPU.}
    \end{tabular}
        
    \caption{Number of epochs and seconds per epoch, averaged across five runs, needed to train the two different approaches on the Stanford cars \cite{stanfordCars} and FGVC-Aircraft \cite{aircraftDataset} datasets.}
    \label{tab:epochs}
\end{table*}

\subsubsection{Experiment 2}

Table \ref{tab:resultados_inception_dense} shows the results obtained when applying our training scheme to the other two backbone architectures selected: InceptionV3 and DenseNet. According to the figures, our approach outperforms the corresponding baseline for both datasets regardless of backbone considered. While analyzing the behavior of our training scheme when using InceptionV3, we can observe that it achieves an improvement of more than 1 percent for the four performance measures. In terms of accuracy, this improvement over the baseline is of 1.3 percent on the Stanford cars dataset and 1.5 percent on the FGVC-Aircraft dataset. Regarding the DenseNet backbone, the improvement with respect to the baseline is about 1.1 percent for all the metrics on both datasets.

\begin{table*}[htb]
    \centering
    \begin{tabular}{l|c|c||c|c}
    \hline  
    \multicolumn{5}{c}{Stanford cars} \\ \hline 
    & FT-InceptionV3 & 0-occlusion-InceptionV3 & FT-DenseNet & 0-occl-DenseNet \\ \hline 
     
    Accuracy & $0.778 \pm 0.023$ & \boldmath$ 0.791 \pm 0.020$ & $0.883\pm0.010$  & \boldmath$0.894 \pm 0.011$ \\	 		
    Precision & $0.788 \pm 0.021 $& \boldmath$0.798 \pm 0.020$ & $0.888\pm0.009$ & \boldmath$0.898 \pm 0.011$ \\ 
    Recall & $0.777 \pm 0.023$& \boldmath$0.791 \pm 0.020$ & $0.882 \pm 0.010$ & \boldmath$0.893 \pm 0.012 $ \\
    F1 & $0.776 \pm 0.023 $& \boldmath$0.790 \pm 0.021$ & $0.882 \pm 0.010 $ & \boldmath$0.893 \pm0.012$ \\ \hline \hline
    
    \multicolumn{5}{c}{FGVC-Aircraft} \\ \hline   
    & FT-InceptionV3 & 0-occlusion-InceptionV3 & FT-DenseNet & 0-occl-DenseNet \\ \hline 
     
    Accuracy & $0.618 \pm 0.029$& \boldmath$0.633 \pm 0.026$ & $0.767 \pm 0.026$ & \boldmath$0.780 \pm 0.025$ \\
    Precision & $0.630 \pm 0.030$ & \boldmath$0.641 \pm 0.029 $ & $0.786 \pm 0.024$ & \boldmath$0.794 \pm 0.023$ \\ 
    Recall & $0.618 \pm 0.028 $& \boldmath$0.633 \pm 0.026$ & $0.767 \pm 0.026$& \boldmath$0.780 \pm 0.025 $ \\ 
    F1 & $0.616 \pm 0.029$& \boldmath$0.630 \pm 0.026$ & $0.768 \pm 0.026$& \boldmath$0.780 \pm 0.025 $\\ \hline  
    \end{tabular}
        
    \caption{Classification performance, averaged across five runs, making use of different backbones on the Stanford cars \cite{stanfordCars} and FGVC-Aircraft \cite{aircraftDataset} datasets. Best results are in bold.}
    \label{tab:resultados_inception_dense}
\end{table*}

\section{Case study}\label{section:case_study}

This section describes an application of the proposed method to a real-world scenario. In particular, we consider the task of food-related scene classification in egocentric images, as detailed below.

\subsection{Dataset}

We evaluated our proposed method on the EgoFoodPlaces dataset \cite{talavera_dataset}, which is composed of 33,810 egocentric images gathered by 11 users and organized in 15 food-related scene classes. By making use of a wearable camera\footnote{\url{http://getnarrative.com/}}, the users regularly recorded an amount of approximately 1,000 images per day. The camera movements and the wide range of different situations that the users experienced during their days, lead to challenges such as background scene variation or changes in lighting conditions. The dataset was manually labeled into a total of 15 different food-related scene classes namely, \textit{bakery shop, bar, beer hall, cafeteria, coffee shop, dining room, food court, ice cream parlor, kitchen, market indoor, market outdoor, picnic area, pub indoor, restaurant, and supermarket.} Table \ref{tab:class_distribution} depicts the distribution of images among the collected classes, with a great imbalance between them.

\begin{table}[htb]
\Large
\resizebox{1.0\columnwidth}{!}{
\begin{tabular}{l|ccccccccccccccc|c}
Class & \rotatebox[origin=c]{90}{Bakery shop} &
\rotatebox[origin=c]{90}{Bar} & 
\rotatebox[origin=c]{90}{Beer hall} &
\rotatebox[origin=c]{90}{Cafeteria} &
\rotatebox[origin=c]{90}{Coffee shop} & 
\rotatebox[origin=c]{90}{Dining room} & 
\rotatebox[origin=c]{90}{Food court} & 
\rotatebox[origin=c]{90}{Ice cream parlor} & 
\rotatebox[origin=c]{90}{Kitchen} & 
\rotatebox[origin=c]{90}{Market indoor} & 
\rotatebox[origin=c]{90}{Market outdoor} & 
\rotatebox[origin=c]{90}{Picnic area} & 
\rotatebox[origin=c]{90}{Pub indoor} &
\rotatebox[origin=c]{90}{Restaurant} & 
\rotatebox[origin=c]{90}{Supermarket} &
\rotatebox[origin=c]{90}{Total} \\ \hline

\begin{tabular}[c]{@{}l@{}}\#Images\end{tabular} & 144 & 1632 & 672 & 1689 & 2313 & 3639 & 204 & 107 & 3837 & 1181 & 1388 & 921 & 511 & 10310 & 5262 & \textbf{33810}\\ \hline 
\end{tabular}}
\caption{Distribution of images per class in the EgoFoodPlaces dataset \cite{talavera_dataset}.}
\label{tab:class_distribution}
\end{table}

\subsection{Experimental results}

This section describes the results obtained when addressing the task of food-related scene classification with our proposed training scheme.

The implementation details are the ones described in Section \ref{sec:exp_implementation} with two exceptions: (1) the resolution of the input images, which in this case is $250 \times 250$ as in \cite{talavera_dataset}; and (2) the application of class oversampling to the fourth largest class (i.e., \textit{dining room}) in order to alleviate the imbalance problem.

Concerning the experimentation, we used the split described in \cite{talavera_dataset}, which includes a division into events for the training and evaluation phases, to make sure that there are no images from the same scene/event in both phases. The validation procedure, in this case, consisted of three partitions, with the following distribution: training set (70\%), validation set (10\%), and test set (20\%). Then, the model was trained and evaluated on the test set. This validation procedure was repeated five times. We report the average performance and the standard deviation calculated across the five runs.

Finally, we considered the four performance metrics detailed in Section \ref{sec:exp_evaluation}: accuracy, precision, recall, and F1 score. Note that, for the per-class metrics (precision, recall, and F1), we calculated the macro- and weighted-averages, as suggested in \cite{talavera_dataset}: \textit{macro} gives equal weight to all classes, while \textit{weighted} is sensitive to imbalances. It is worth noting the relevance of these two average values due to the unbalanced nature of the dataset.

\subsubsection{Classification performance}

For the evaluation of our proposal, we followed the experimental setup described in Section \ref{sec:exp_setup}, but using the EgoFoodPlaces dataset to train the ResNet50 architecture with the classical procedure (FT-ResNet50) and with our training scheme (0-occlusion). Additionally, we compared our results with the ones reported in \cite{talavera_dataset}, the state-of-the-art approach for this dataset.

Table \ref{tab:resultados_egocentric} reports the results obtained for the different approaches. As can be seen, training a ResNet50 with our proposed scheme (0-occlusion) allows us to achieve a higher accuracy than the one obtained with the baseline (FT-ResNet50). Moreover, the proposed method also achieves higher weighted averages for the other three metrics considered (precision, recall, and F1). It is worth noting that, due to the high imbalance of the dataset, the weighted metrics are more informative than the macro values. Concerning the latter, the differences between both methods are minimal, with the same values for precision and F1, and a slightly higher macro recall in favor of the baseline.

\begin{table*}[htb]
    \centering
   \begin{tabular}{l|c|c|c}
    
   & Hierarchical approach \cite{talavera_dataset} & FT-ResNet50   & 0-occlusion \\ \hline
        
    Macro Precision & 0.56 & \boldmath$0.59 \pm 0.03$ & \boldmath$0.59  \pm 0.05$   \\ 
    Macro Recall & 0.53 & \boldmath$0.55 \pm 0.03$ &  $0.54 \pm 0.06$ \\
    Macro F1 & \textbf{0.53} & \boldmath$0.53 \pm 0.04$ & \boldmath$0.53 \pm 0.06$ \\ \hline
    
    Weighted Precision & 0.65 & $0.67 \pm 0.02$ & \boldmath$0.68 \pm 0.03$ \\
    Weighted Recall & \textbf{0.68} & $0.67 \pm 0.03$ & \boldmath$0.68 \pm 0.04$ \\ 
    Weighted F1 & 0.65 & $ 0.64 \pm 0.03 $&  \boldmath$0.66 \pm 0.04 $ \\ \hline
    
    Accuracy & \textbf{0.68} & $0.67 \pm 0.03$ & \boldmath$0.68 \pm 0.04$ \\ \hline
    \end{tabular}
    
    \caption{Classification performance, averaged across five runs, of the different approaches on the EgoFoodPlaces dataset \cite{talavera_dataset}. Best results are in bold.}
    \label{tab:resultados_egocentric}
\end{table*}

    
    
    

If we analyze the results achieved by the state-of-the-art \cite{talavera_dataset} and compare them with the proposed method, we can see that our approach achieves better results in four out of the seven performance measures, whereas the remaining three are equal. We find important to point out that our approach makes use of only just one classifier (ResNet50), while the model presented in \cite{talavera_dataset} uses a hierarchical ensemble composed of six VGG16 networks. Therefore, the complexity of our model is significantly lower, not only because we have one single classifier but also because our backbone model ResNet50 has a lower number of parameters than their VGG16 networks. Therefore, we can conclude that our proposed method is able to achieve similar performance results with a less complex architecture and a computationally less expensive approach.

Finally, the impact of the different approaches on the individual classes is presented in Table \ref{tab:resultados_egocentric_class}. As can be seen, our method (0-occlusion) shows a behavior very similar to the baseline approach (FT-ResNet50), with slightly higher rates in seven classes and three ties. Analyzing the figures obtained with the hierarchical approach \cite{talavera_dataset}, our method achieves better results in eight classes. More specifically, the results in which our approach outperforms the state-of-the-art correspond to the four most represented classes (\textit{restaurant}, \textit{supermarket}, \textit{kitchen}, and \textit{dining room}). Also noteworthy is the improvement achieved for the class \textit{food court}, which could not be classified by the hierarchical model (true positive rate of 0.00). However, there are five classes for which the hierarchical model gets a better performance, including \textit{beer hall}, \textit{cafeteria}, and \textit{coffee shop}. We deduce that this is due to the benefits of classifying them in a hierarchical fashion.

\begin{table}[htb]
\Large
\resizebox{1.0\columnwidth}{!}{
\begin{tabular}{l|ccccccccccccccc}
Class & \rotatebox[origin=c]{90}{Bakery shop} &
\rotatebox[origin=c]{90}{Bar} & 
\rotatebox[origin=c]{90}{Beer hall} &
\rotatebox[origin=c]{90}{Cafeteria} &
\rotatebox[origin=c]{90}{Coffee shop} & 
\rotatebox[origin=c]{90}{Dining room} & 
\rotatebox[origin=c]{90}{Food court} & 
\rotatebox[origin=c]{90}{Ice cream parlor} & 
\rotatebox[origin=c]{90}{Kitchen} & 
\rotatebox[origin=c]{90}{Market indoor} & 
\rotatebox[origin=c]{90}{Market outdoor} & 
\rotatebox[origin=c]{90}{Picnic area} & 
\rotatebox[origin=c]{90}{Pub indoor} &
\rotatebox[origin=c]{90}{Restaurant} & 
\rotatebox[origin=c]{90}{Supermarket} \\ \hline

\begin{tabular}[c]{@{}l@{}}Hierarchical app. \cite{talavera_dataset}\end{tabular} & 0.39 & 0.31 & \textbf{0.89} & \textbf{0.45} & \textbf{0.59} & 0.58 & 0.00 & \textbf{0.52} & 0.89 & \textbf{0.70} & 0.28 & 0.00 & \textbf{0.85} & 0.70 & 0.85\\ 

\begin{tabular}[c]{@{}l@{}}FT-ResNet50\end{tabular} & \textbf{0.63} & 0.31 & 0.24 & 0.38 & 0.49 & 0.66 & \textbf{0.53} & 0.50 & 0.89 & 0.60 & \textbf{0.57} & 0.00 & 0.78 & \textbf{0.73} & 0.90 \\  

\begin{tabular}[c]{@{}l@{}}0-occlusion\end{tabular} & 0.60 & \textbf{0.32} & 0.26 & 0.35 & 0.48 & \textbf{0.72} & 0.43 & \textbf{0.52} & \textbf{0.90} & 0.60 & 0.53 & 0.00 & 0.80 & \textbf{0.73} & \textbf{0.92} \\ \hline 
\end{tabular}}

\caption{True positive rate per class, averaged across five runs, of the different approaches on the EgoFoodPlaces dataset \cite{talavera_dataset}. Best results are in bold.}
\label{tab:resultados_egocentric_class}
\end{table}

Going deeper into the results obtained and given the characteristics of the EgoFoodPlaces dataset, we can draw some additional conclusions. Firstly, we can observe that the classification improves when using our approach for (1) classes where the scene to recognize is right in front of the camera users (e.g., \textit{restaurant}), and (2) classes that tend to share descriptors even if recorded at different locations (e.g., \textit{dining room} or \textit{supermarket}). 
Those results inherit that the model is able to learn the relevant features in the scene when it is self-contained, which is closely related to the fine-grained datasets evaluated in Section \ref{section:experimental_results}.

Analyzing the images we can also see that, in some classes (e.g., \textit{food court}, \textit{cafeteria}, \textit{market outdoor}), there is more background than foreground information necessary for the identification of the scene; that is, the image is composed of characteristics that an observer would not find relevant for the distinction of an event. Therefore, the main difficulty in learning these scenes is that not only the locations vary but also they are composed of elements common to other scenes. In this case, including other relevant regions along with a limited amount of samples available per class might represent imposed noise and lead to a lower performance in our approach compared to the baseline. This issue could be addressed with the extension of the dataset. 

\subsubsection{Model inspection}

We analyzed not only the classification performance of our training scheme but also its ability to make predictions. In particular, we aimed to find out if the proposed approach is able to improve the robustness of a CNN classifier and make it sensible to more features. For this reason, we carried out two additional experiments: (1) we analyzed the behavior of the models making use of a visual explanation algorithm, and (2) we randomly erased some areas of the test images before evaluating the models on them.

\begin{figure} 
    \centering
    \begin{tabular}{ccc}
     
     \begin{subfigure}[b]{0.275\linewidth}
  \includegraphics[width=2.75cm, height=2.75cm]{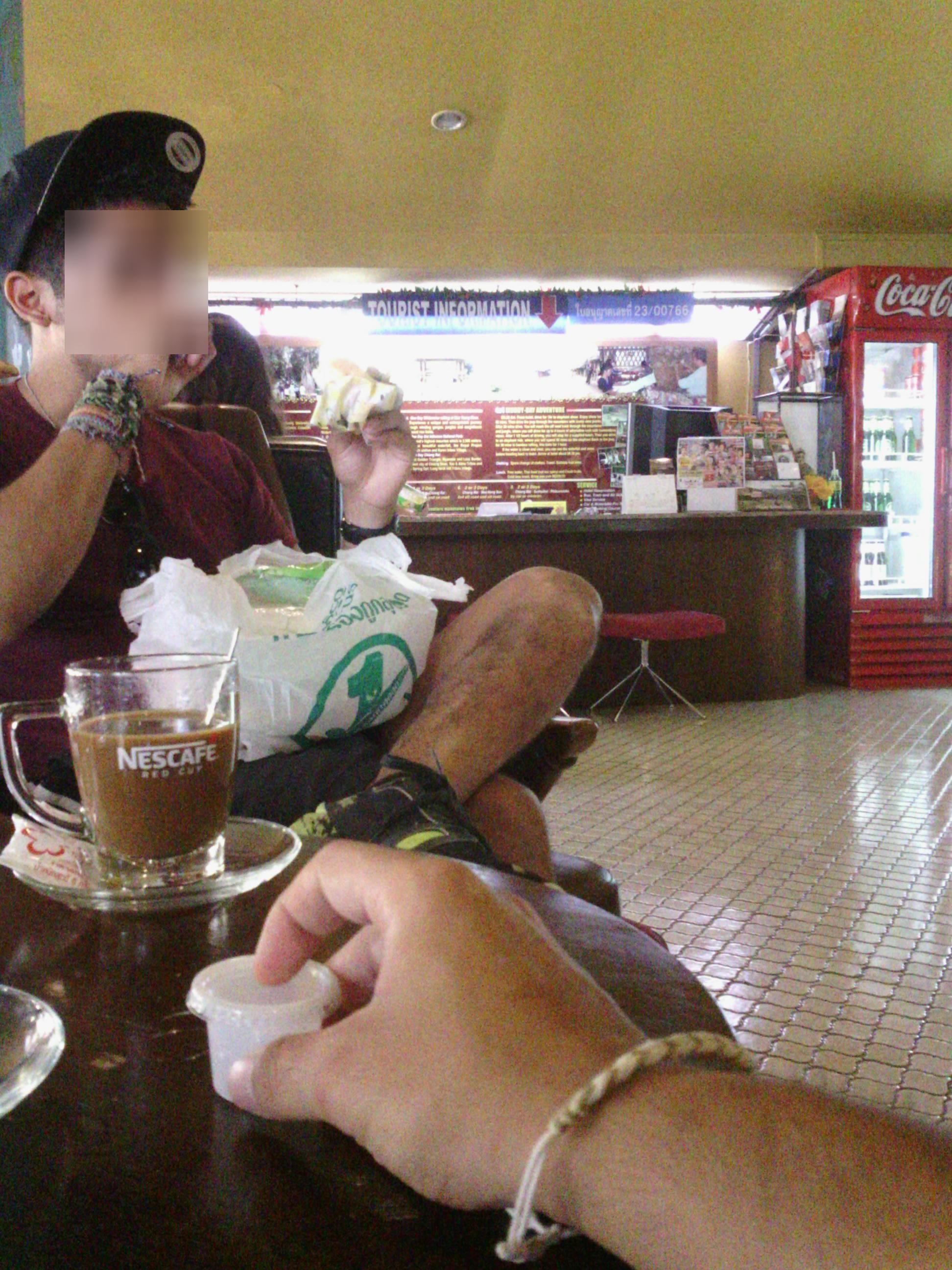}
     \end{subfigure}
     &
     \begin{subfigure}[b]{0.275\linewidth}
  \includegraphics[width=2.75cm, height=2.75cm]{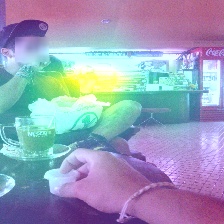}
     \end{subfigure}
     &
     \begin{subfigure}[b]{0.275\linewidth}
  \includegraphics[width=2.75cm, height=2.75cm]{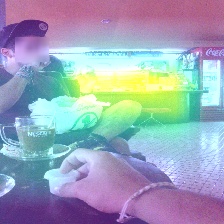}
     \end{subfigure}
     \\
     
     \begin{subfigure}[b]{0.275\linewidth}
  \includegraphics[width=2.75cm, height=2.75cm]{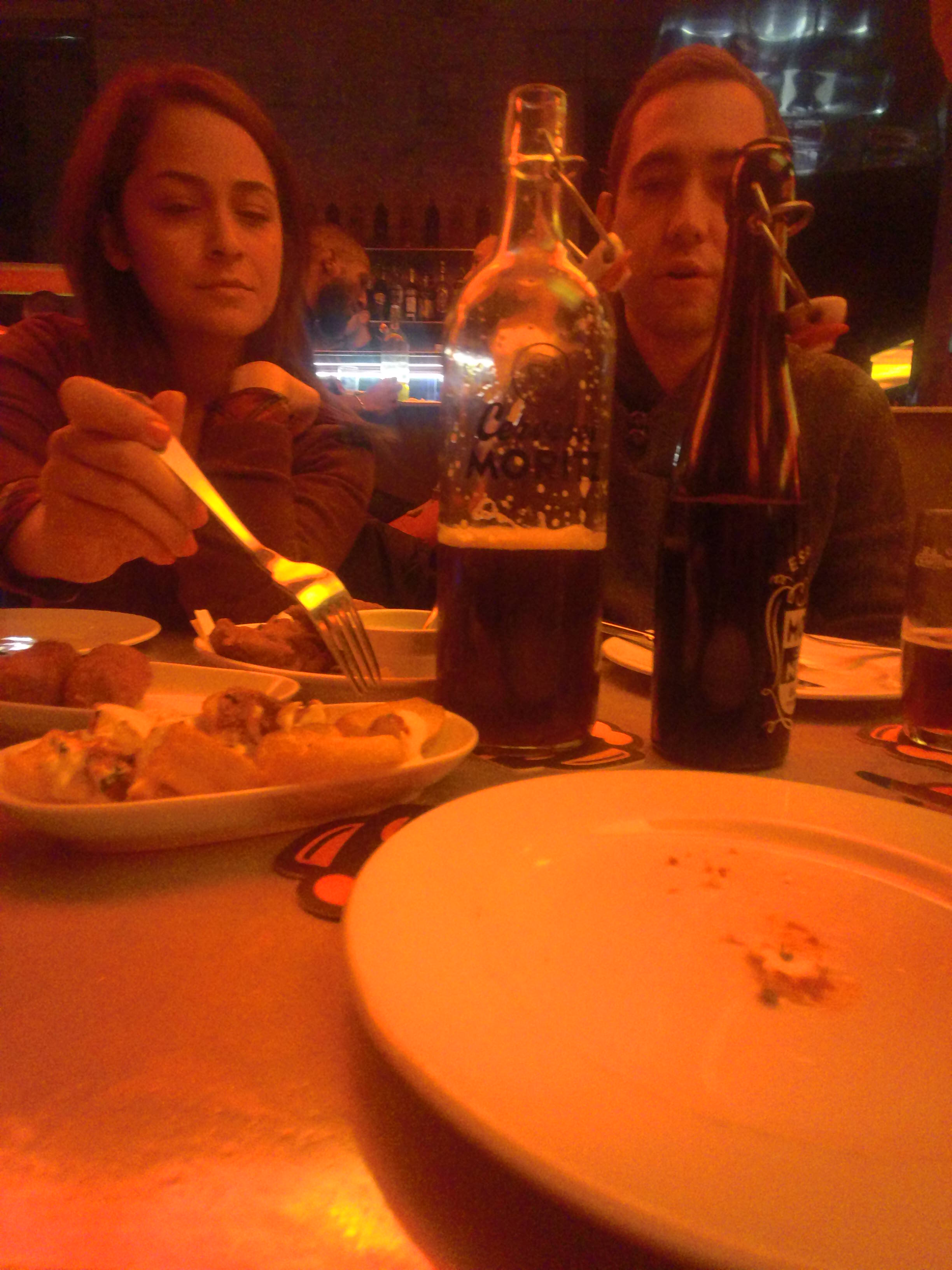}
     \end{subfigure}
     &
     \begin{subfigure}[b]{0.275\linewidth}
  \includegraphics[width=2.75cm, height=2.75cm]{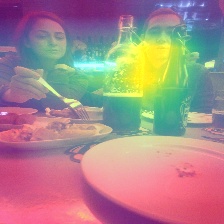}
     \end{subfigure}
    &
     \begin{subfigure}[b]{0.275\linewidth}
  \includegraphics[width=2.75cm, height=2.75cm]{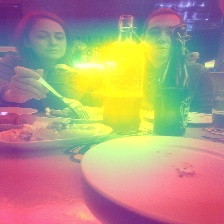}
     \end{subfigure}
     \\
     
     \begin{subfigure}[b]{0.275\linewidth}
  \includegraphics[width=2.75cm, height=2.75cm]{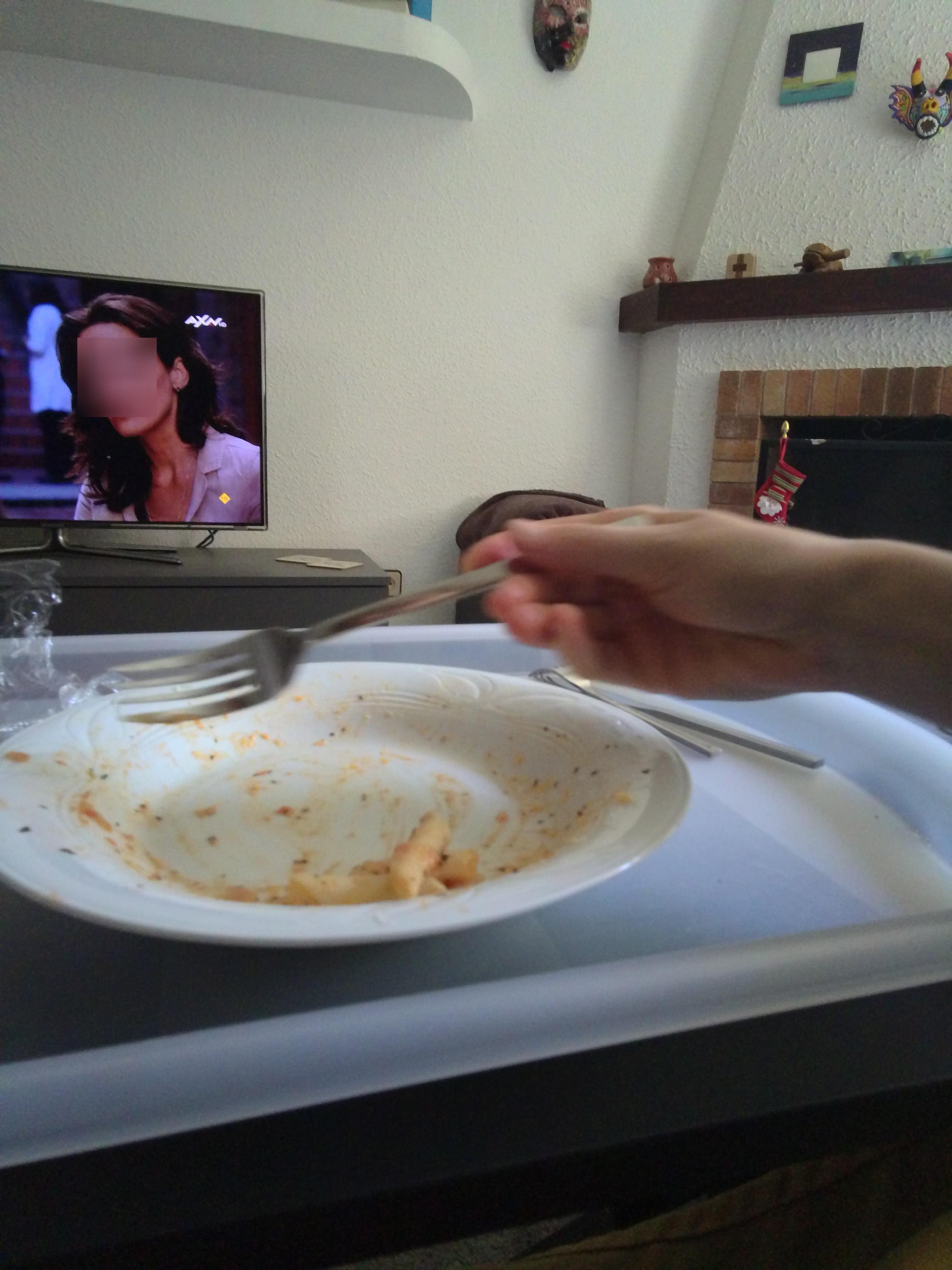}
     \end{subfigure}
     &
     \begin{subfigure}[b]{0.275\linewidth}
  \includegraphics[width=2.75cm, height=2.75cm]{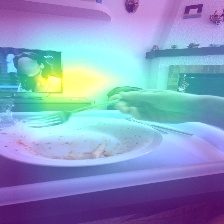}
     \end{subfigure}
     &
     \begin{subfigure}[b]{0.275\linewidth}
  \includegraphics[width=2.75cm, height=2.75cm]{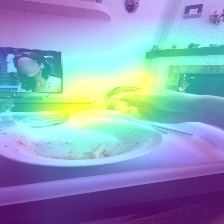}
     \end{subfigure}
     \\
     
     \begin{subfigure}[b]{0.275\linewidth}
  \includegraphics[width=2.75cm, height=2.75cm]{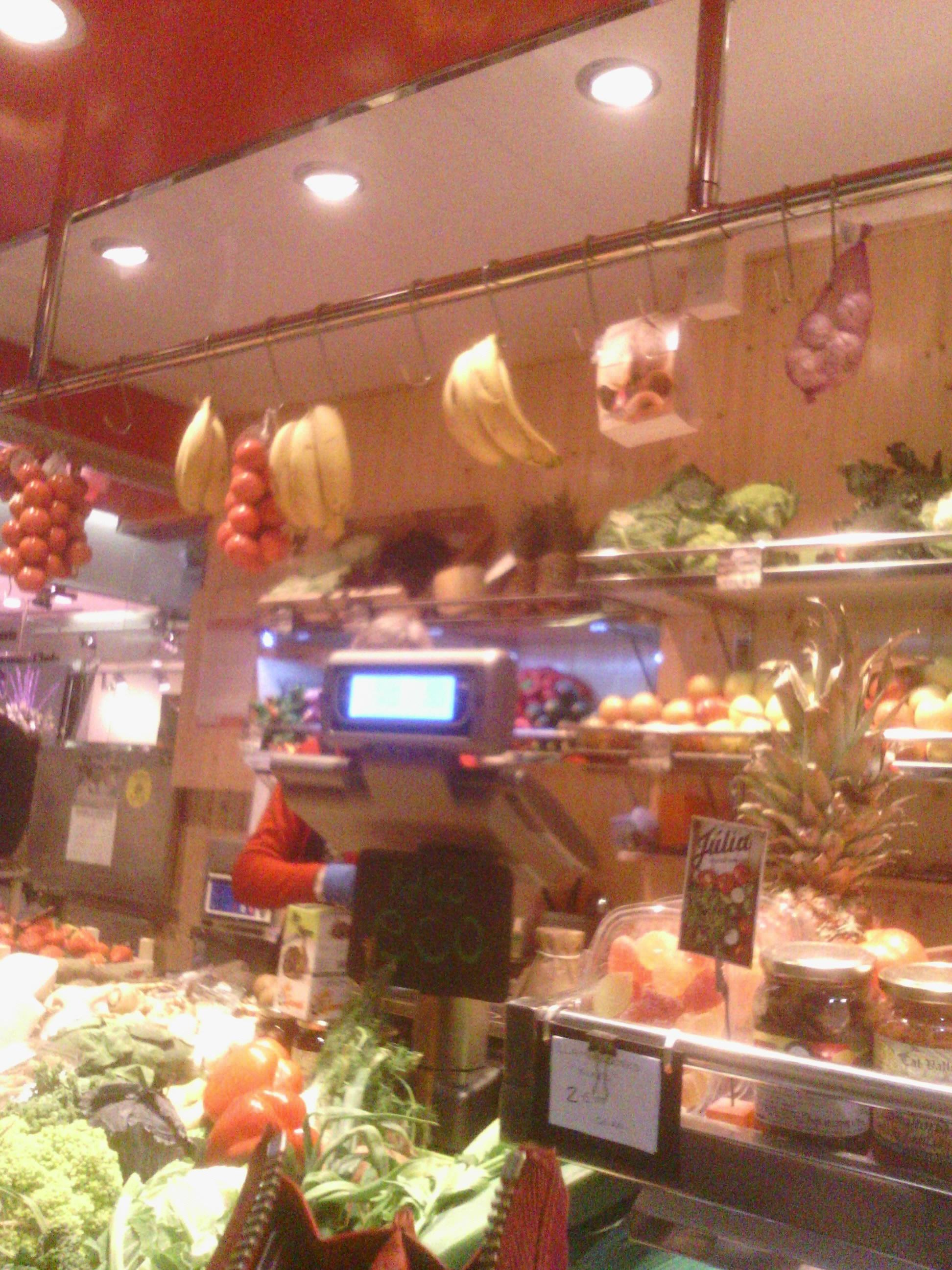}
     \end{subfigure}
     &
 \begin{subfigure}[b]{0.275\linewidth}
  \includegraphics[width=2.75cm, height=2.75cm]{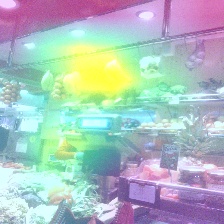}
     \end{subfigure}
     &
     \begin{subfigure}[b]{0.275\linewidth}
  \includegraphics[width=2.75cm, height=2.75cm]{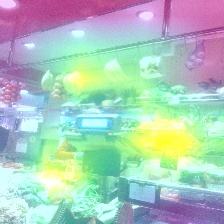}
     \end{subfigure}
     \\
     
     \begin{subfigure}[b]{0.275\linewidth}
  \includegraphics[width=2.75cm, height=2.75cm]{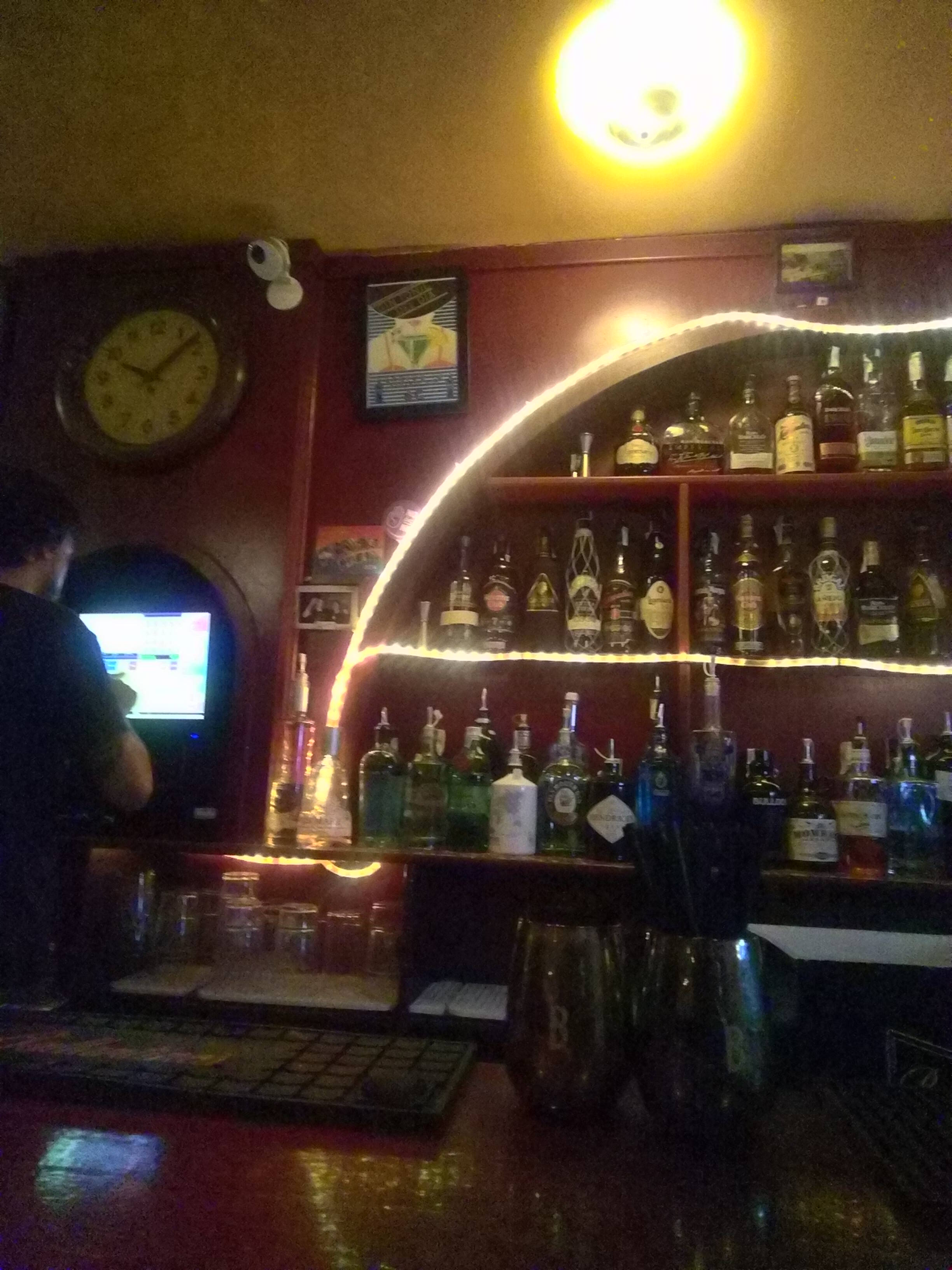}
     \end{subfigure}
     &
     \begin{subfigure}[b]{0.275\linewidth}
  \includegraphics[width=2.75cm, height=2.75cm]{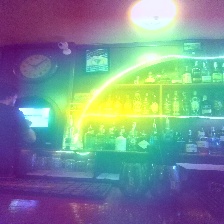}
     \end{subfigure}
     &
     \begin{subfigure}[b]{0.275\linewidth}
  \includegraphics[width=2.75cm, height=2.75cm]{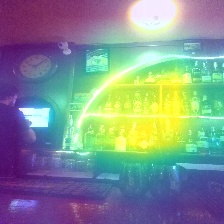}
     \end{subfigure}
  \\
  
     \begin{subfigure}[b]{0.275\linewidth}
  \includegraphics[width=2.75cm, height=2.75cm]{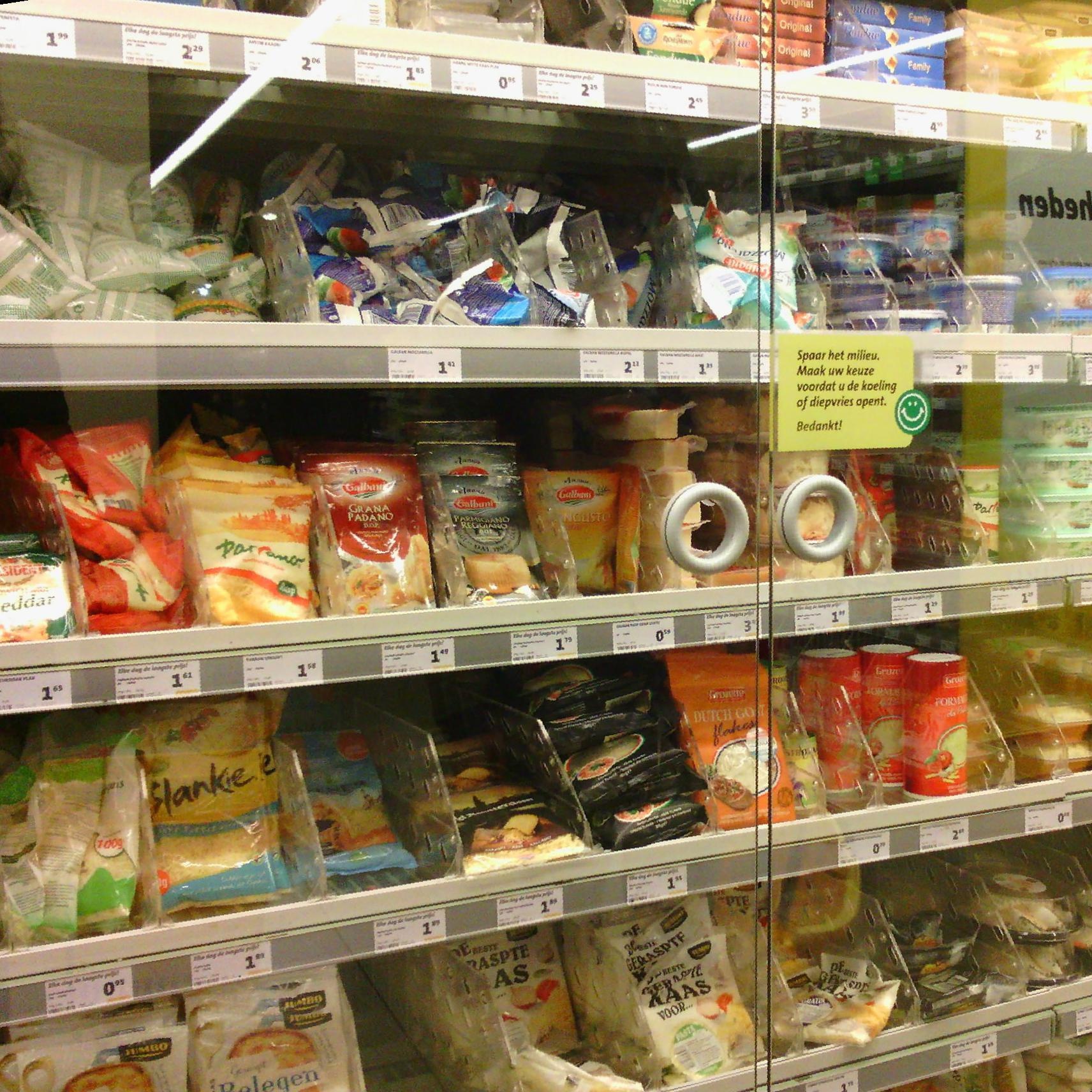}
  \caption{}
   \end{subfigure}
   &
     \begin{subfigure}[b]{0.275\linewidth}
  \includegraphics[width=2.75cm, height=2.75cm]{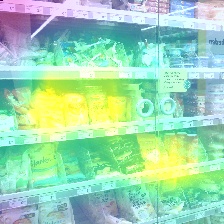}
  \caption{}
    \end{subfigure}
     &
     \begin{subfigure}[b]{0.275\linewidth}
  \includegraphics[width=2.75cm, height=2.75cm]{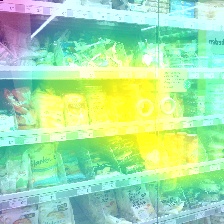}
  \caption{}
    \end{subfigure}
    
    \end{tabular}
    \caption{(a) Input images, (b) heat maps generated by Grad-CAM for the baseline FT-ResNet50, and (c) heat maps generated by Grad-CAM for the model trained with the proposed training scheme (0-occlusion).} 
    \label{fig:analysis_ego}
\end{figure}

In the first experiment, our target was to demonstrate that the regions considered as relevant by the trained models were more and bigger when applying our training scheme than when following the classical procedure. For this purpose, we applied the Grad-CAM algorithm to the last convolutional layer of the two ResNet50 models previously trained on the EgoFoodPlaces dataset: one trained using the classical procedure (FT-ResNet50), and the other one using our training scheme with 0-occlusion. As a result, we obtained the heat maps that allow us to visualize the regions that are important to the models when making a prediction for a given image. Figure \ref{fig:analysis_ego} depicts some representative images along with their corresponding heat maps for each model. As can be observed, our model took into account bigger regions than the baseline method (FT-ResNet50) when processing the same target images. Besides, it can be seen that the model trained with our proposed method bases its decisions on more regions than when using the classical procedure. Furthermore, the regions that the baseline model took into account when making a decision were also taken into account by the proposed model. This demonstrates that when using the proposed training scheme, the model does not forget the learned features, but just learns to recognize other features.

Finally, we conducted the second experiment to test the robustness of our training scheme. For this purpose, we hid some regions of the test images by randomly erasing them, as proposed in \cite{randomErasing}. After that, we compared how the two approaches (FT-ResNet50 and 0-occlusion) performed on the modified test set. Table \ref{tab:resultados_ocultando_test} presents the results for this experiment. As can be observed, the proposed approach (0-occlusion) performs better than the baseline model (FT-ResNet50). This means that our model does not suffer as much when some areas of the image are erased or hidden, demonstrating its robustness. It is also worth noting that these results are consistent with the ones obtained in the previous experiment, and demonstrate that our model makes use of more and bigger regions than the baseline approach to make a prediction for a target image.

\begin{table}[htb]
    \centering
    \begin{tabular}{l|c|c}
    & FT-ResNet50 & 0-occlusion  \\\hline
    Macro Precision & $0.53 \pm 0.01 $ & \boldmath$0.54 \pm 0.02 $   \\ 
    Macro Recall & $0.47 \pm 0.02 $ & \boldmath$0.48 \pm 0.03$ \\
    Macro F1 & $0.47 \pm 0.02 $& \boldmath$0.48 \pm 0.05$ \\ \hline
    Weighted Precision & \boldmath$0.63 \pm 0.02$ &   \boldmath$0.63 \pm 0.03$ \\
    Weighted Recall & $0.59 \pm 0.02 $& \boldmath$0.65 \pm 0.03$ \\
    Weighted F1 & \boldmath$0.59 \pm 0.02 $  &   \boldmath$0.59 \pm 0.02$ \\ \hline
    Accuracy & $0.59 \pm 0.02$ & \boldmath$0.60 \pm 0.02$ \\
    \hline
    \end{tabular}
    \caption{Classification performance, averaged across five runs, of the baseline method and the proposed training scheme when we randomly hid some regions on the test images. Best results are in bold.}
    \label{tab:resultados_ocultando_test}
\end{table}

\section{Conclusion}\label{section:conclusion}

This research work presents a novel training scheme that improves the robustness and generalization ability of CNNs applied to image classification. The idea is to force the model to learn as many features as possible when making a class selection. For this purpose, we apply a visual explanation algorithm to identify the areas on which the model bases its decisions. After identifying those areas, we occluded them and trained the model with a combination of the modified images and the original ones. 
In this manner, the model is not able to base its prediction on the occluded regions and is forced to use other areas. Consequently, the model also learns to pay attention to those regions of the target image that, \textit{a priori}, are not so informative for its classification.

To evaluate the proposed method, we carried out different experiments on two popular datasets used for fine-grained recognition tasks: Stanford cars and FGVC-Aircraft. From the obtained results, we can confirm our initial hypothesis: our method forces the network to learn additional features that help it distinguish between very similar classes, showing its suitability for fine-grained classification problems. More specifically, and within the different evaluated configurations, the 0-occlusion approach has shown to be the most appropriate setting. Furthermore, we demonstrated the adequacy of our training scheme regardless of the backbone architecture considered.

We further experimented with a real-case study focused on the classification of food-related scenes. We analyzed the impact of our training scheme by comparing it with a baseline method and, to the best of our knowledge, with the state-of-the-art approach that follows an ensemble composed of six CNNs \cite{talavera_dataset}. The results achieved with our method were comparable or even better than the ones obtained with the state-of-the art approach despite making use of just one network, thus reducing the level of complexity while maintaining a competitive performance. Furthermore, our method is computationally less expensive, as the chosen backbone (ResNet50) has fewer parameters than the VGG16 used in \cite{talavera_dataset}. Finally, we carried out several occlusion and visual explanation experiments, showing that our method improves the robustness of the classifier by forcing it to base its decisions on more features.

As a future line of research, it would be interesting to apply the same methodology not only to input images but also at different convolutional levels, as it is usually done with the regularization technique known as dropout. In other words, the feature maps obtained at different levels could be analyzed and occluded in the same way that we did with the input images. This idea would force the model to pay attention to different characteristics on the feature maps, thereby improving the robustness of the model at different levels of the learning process. 

\begin{acknowledgements}
We would like to thank the Center for Information Technology of the University of Groningen for their support and for providing access to the Peregrine high performance computing cluster.
\end{acknowledgements}

%
\section*{Conflict of interest}
The authors declare that they have no conflict of interest.

\bibliographystyle{spphys}  
\bibliography{biblio}   

%
%

\end{document}